\RequirePackage{fix-cm}
\documentclass[smallextended]{svjour3}       
\smartqed  
\usepackage{graphicx}
\usepackage{hyperref}
\usepackage{caption}
\usepackage{subcaption}
\usepackage{graphicx}
\usepackage{amsmath}
\usepackage{algorithm}
\usepackage[noend]{algpseudocode}
\usepackage{amssymb}
\usepackage{latexsym}

\journalname{\tiny{This is a pre-print of an article published in Pattern Analysis and Applications.}}
\begin{document}

\title{Bag of Recurrence Patterns Representation for Time-Series Classification
}


\author{Nima Hatami         \and
        Yann Gavet 			\and
        Johan Debayle
}

\institute{N. Hatami, Y. Gavet, J. Debayle \at
              The SPIN center, \'Ecole Nationale Sup\'erieure des Mines de Saint-\'Etienne, France \\
              Tel.: +33 472437543\\
              \email{hatami@creatis.insa-lyon.fr}            \\
             N. Hatami is currently with the CREATIS center, INSA-Lyon and University of Lyon, France}  

\date{Received: date / Accepted: date}

\maketitle

\begin{abstract}

Time-Series Classification (TSC) has attracted a lot of attention in pattern recognition, because wide range of applications from different domains such as finance and health informatics deal with time-series signals. Bag of Features (BoF) model has achieved a great success in TSC task by summarizing signals according to the frequencies of "feature words" of a data-learned dictionary. This paper proposes embedding the Recurrence Plots (RP), a visualization technique for analysis of dynamic systems, in the BoF model for TSC. While the traditional BoF approach extracts features from 1D signal segments, this paper uses the RP to transform time-series into 2D texture images and then applies the BoF on them. Image representation of time-series enables us to explore different visual descriptors that are not available for 1D signals and to treats TSC task as a texture recognition problem. Experimental results on the UCI time-series classification archive demonstrates a significant accuracy boost by the proposed Bag of Recurrence patterns (BoR), compared not only to the existing BoF models, but also to the state-of-the art algorithms.

\end{abstract}

\section{Introduction}
\label{intro}
A time-series is a sequence of data points (measurements) which has a natural temporal ordering. Many important real-world pattern recognition tasks deal with time-series analysis. Biomedical signals (e.g. EEG and ECG), financial data (e.g. stock market and currency exchange rates), industrial devices (e.g. gas sensors and laser excitation), biometrics (e.g. voice, signature and gesture), video processing, music mining, forecasting and weather are examples of application domains with time-series nature \cite{Ref19,Ref24,Ref23,Ref5,Ref30}. The time-series mining motivations and tasks are mainly divided into curve fitting, function approximation, prediction and forecasting, segmentation, classification and clustering categories. In a univariate time-series classification, $x^{n} \rightarrow y^{n}$ so that $n$-th series of length $l$: $x^{n} = (x^{n}_{1}, x^{n}_{2}, ..., x^{n}_{l})$ is associated with a class label $y^{n} \in \{1,2,..,N_{c}\}$. It is worth noting that although this paper is mainly focused on time-series classification problem, the proposed method can be easily adapted to the other tasks such as clustering and anomaly detection.

The existing time-series classification methods may be categorized from different perspectives. Regarding the feature types, "frequency-domain" methods include spectral analysis and wavelet analysis; while "time-domain" methods include auto-correlation, auto-regression and cross-correlation analysis. Regarding the classification strategy, it can also be divided into "instance-based" and "feature-based" methods. The former measures similarity between any incoming test sample and the training set; and assigns a label to the most similar class (the euclidean distance based 1- Nearest Neighbor and Dynamic Time Wrapping (DTW) are two popular and widely used methods of this category \cite{Ref6,Ref7,Ref8,Ref9,Ref10}). The latter first transforms the time-series into the new space and extract more discriminative and representative features in order to be used by a pattern classifier which aiming to find optimum classification boundaries \cite{Ref11,Ref12,Ref13,Ref14,Ref16}.
\begin{figure*}[t!]
\centering
\includegraphics[scale=.35]{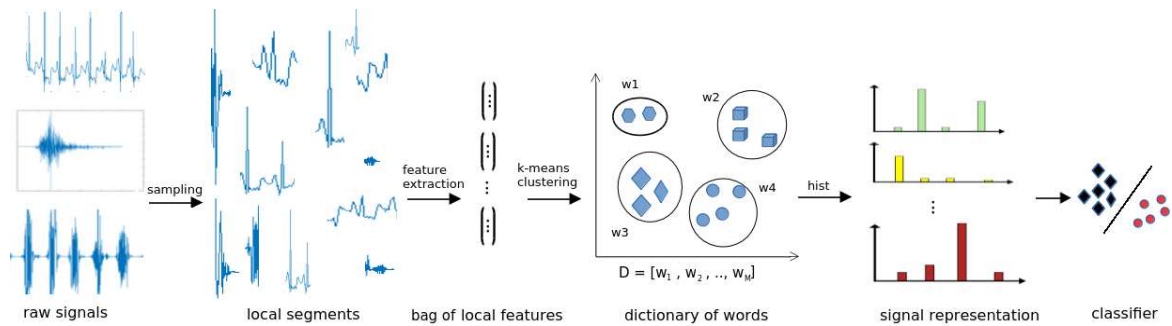}
\caption{The traditional BoF model for time-series classification.}
\label{fig:figure00}
\end{figure*}

The BoF (also known as bag-of-words) is a simplifying representation that originally used in natural language processing and information retrieval. In this model, a text (such as a sentence or a document) is represented as the bag (multiset) of its words, disregarding grammar and even word order but keeping multiplicity. The BoF is commonly used for document classification where the (frequency of) occurrence of each word is used as a feature for training a classifier. Later, the BoF model has also adopted for computer vision and signal processing.
Recently, BoF model has achieved a high recognition rate for time-series classification \cite{Ref18,Ref19,Ref20,Ref25,Ref31}. The main idea is to represent a signal by a histogram of “feature words” of a data-learned dictionary (codebook). There are five steps to perform BoF model: i) creating local segments from the input signals, ii) extracting local features from the segments and creating the "bag", iii) learning codebook, iv) quantizing features using codebook and representing samples by histogram of words, and v) classification. Fig.\ref{fig:figure00} shows the traditional BoF model for time-series classification.\\
Traditionally, when applying BoF on a series, features are extracted from "bag of segments" obtained by sliding a window over the signal. On the way towards exploring different discriminative and informative features, the proposed work aims to investigate an alternative approach devoted for time-series analysis, by transforming 1D signals into 2D images before applying BoF. This provides extra tools for addressing time-series analysis by taking advantage of existing advanced image processing and computer vision methods. Beside the currently available signal processing measures and statistics that traditionally being used for time-series feature extraction, the proposed approach provides access to different efficient features and descriptors such as Scale-invariant feature transform (SIFT)\cite{Ref39}, Histogram of Oriented Gradients (HOG) \cite{Ref44} and Local Binary Patterns (LBP)\cite{Ref40} which are defined only in image space.

There are various approaches for visualization of time-series with the advantage of better understanding and analyzing them \cite{vis_ts}. This paper investigates using a Recurrence Plot (RP) \cite{Eckmann} - an advanced technique of nonlinear data analysis in chaos theory - to visualize time-series. The RP is the visual inspection of higher dimensional phase space trajectories and exhibits global characteristic ($typology$) and local patterns ($texture$). Instead of quantization of RP matrix with measures such as recurrence rate, determinism, mean diagonal line length and entropy as traditionally suggested, Vinicius et al. \cite{RP_svm} propose to treat plots as gray-level texture images and apply an SVM classifier on texture features. In this work instead, we propose a Bag of Recurrence Patterns (BoR) model by applying a popular method of BoF on vocabulary of local image patterns. The experimental results on the publicly available UCI time-series dataset \cite{Ref33} demonstrate that treating the time-series classification problem as image texture recognition boosts the accuracy by extracting more representative and discriminative information using features and descriptors which are only applicable in image space.

\section{Related Work}

This section briefly reviews the recent contributions on time-series classification using BoF model, novel texture image recognition techniques and application of the RP on TSC task.

Recently, there has been a big effort to improve and adapt BoF for time-series application. 
Multiple subsequences selected from random locations and of random lengths suggested by Baydogan et. al. \cite{Ref18} to handle the time warping of patterns. Mel-Frequency Cepstrum Coefficients (MFCC) features are used in \cite{Ref28} for music genre classification and artist identification. Vectors of Discrete Wavelet Transform (DWT) features from each of the local segments is used in BoF framework for EEG and ECG classification \cite{Ref19}. Grabocka et. al. \cite{Ref22} build words using coefficients of  local polynomial approximations and use their histogram to classify repetitive time series. Bailly et al. \cite{Ref25} adapted SIFT image descriptors for 1D domain. A BoF approach using statistical features is applied to identify general motion primitives for modeling different human activities \cite{Ref24}. Bromuri et al. \cite{Ref23} combine BoF and kernel methods-based dimensionality reduction to perform multi-label time-series classification on health records of chronically ill patients. In another research, a method for discovering characteristic patterns in a time series based on Symbolic Aggregate approXimation and Vector Space Model (SAX-VSM) is proposed by \cite{Ref21}. Raw signals are converted into bags of words using SAX; followed by calculation of term frequency-inverse document frequency (TF-IDF). Class labels are assigned using maximal cosine similarity. Wang ans Oates \cite{Ref30} inspired by the max-pooling approaches in Convolutional Neural Networks (CNN) and extend SAX to multivariate synchronous vital signs data. Finally, \cite{Ref31} proposed a time warping SAX in order to integrate the temporal correlation. Instead of sampled randomly or uniformly, \cite{Ref47} used a peak and valley detection algorithm and subsequences are sampled from these peaks and valleys. Then, Histogram of Oriented Gradients (HOG-1D) and Dynamic time warping-Multidimensional scaling (DTW-MDS), are proposed to describe segments. Bag-of-SFA-Symbols (BOSS) ensemble \cite{Ref48} is a noise-resistant method that combines Symbolic Fourier Approximation (SFA) representation and BoF model. Learned Pattern Similarity (LPS)\cite{Ref49} is an unsupervised approach to represent and measure the similarity and models the dependency structure in time series that generalizes the concept of auto-regression to local autopatterns.

Image texture recognition is one of the important tasks in computer vision and image processing. Here, a short review of some texture recognition methods are given. Multiscale rotation-invariant sampling based texture image classification is proposed in \cite{Dong_2017}. A multiscale wavelet transform is used to decompose the magnitude pattern (MP) mapping of a texture and the sampled
directional mean vectors (SDMVs) of each wavelet subband is computed. In addition, frequency vectors (FVs) of the sign pattern mappings are also constructed for capturing the structural information of textures. The texture information at a given scale is described by three SDMVs and an FV.
A texture classification approach is proposed by modeling joint distributions
of local patterns with Gaussian mixture models (GMM) \cite{Lategahn_2010}. A given texture region is filtered by a set of filters and subsequently the joint probability density function is estimated by GMMs. 
A statistical approach to represent the marginal distribution of the wavelet coefficients using finite mixtures of generalized Gaussian (MoGG) distributions is proposed \cite{Allili_2012}. The MoGG captures a wide range of histogram shapes, which provides better description and discrimination of texture. Dense Micro-Block Difference (DMD) \cite{Mehta_2016} local features are proposed for texture images. The local features are then projected into a low dimensional space using Random Projections and the low dimensional features are encoded into a image descriptor using Fisher vectors. In another research, energy features are extracted from shearlet (extension of wavelets) subbands and their dependences are modelled using linear regression \cite{Dong_2015}. The regression residuals are used for defining the distance from a texture image to a class.

Although RP \cite{Eckmann} is proposed a long time ago, application of RP on TSC has not been fully explored. In fact, there are not so many types of approaches, when it comes to application of time-series images on TSC. Traditionally, PRs are described by Recurrence Quantification Analysis (RQA) measures, which quantifies the number and duration of recurrences of a dynamical system presented by its state space trajectory \cite{Zbilut1992,Marwan2002}. Roma et al. \cite{Roma2013} explored the use of the RQA features for auditory scene classification. These features are computed over a thresholded similarity matrix computed from windows of MFCC features. Added to traditional MFCC statistics, they improve accuracy. Their experiments suggest that RQA features have some discriminating power with respect to auditory scenes, that is not captured in basic MFCC statistics. In another application, RQA parameters are extracted from the RP in order to classify the EEG signals into normal, epileptic (ictal) and background (pre-ictal) classes \cite{Acharya2011}. Yang \cite{Yang_2011} and Chen and Yang \cite{Chen_2012} applied RQA on Vectorcardiogram (VCG) and heart rate variability (HRV) signals in multiple wavelet scales and investigated different classifiers for the identification of cardiac disorders (discrimination between Myocardial Infarction (MI)/congestive heart failure (CHF) and healthy control (HC) subjects). The multiscale RQA are found to be adequately discriminatory to capture the dynamic variations between HC/CHF and MI. Silva et al. \cite{silva2013} measured the similarity between RPs using Campana-Keogh (CK-1) distance, a Kolmogorov complexity-based distance that uses video compression algorithms to estimate image similarity. They show that RP allied to CK-1 distance lead to accuracy improvements compared to Euclidean distance and DTW on the UCR data sets. In another work by Silva et al. \cite{RP_svm} image texture features i.e. LBP, Grey Level Co-occurrence Matrix (GLCM), Gabor filters, and Segmentation-based Fractal Texture Analysis (SFTA) are extracted from RPs and then, TSC task is addressed by applying a SVM classifier on combination of those features. Taking advantage of the Convolutional Neural Networks (CNN)'s high performance on image classification, Hatami et al. \cite{hatami-cnn} first transformed time-series signals into texture images (using RP) and then classify them by a 2-layer CNN model. Obtaining high recognition rates on the UCR datasets demonstrates the efficiency of the proposed pipeline using RP features and CNN classifiers on TSC.

\begin{figure*}[t!]
\centering
\includegraphics[scale=.4]{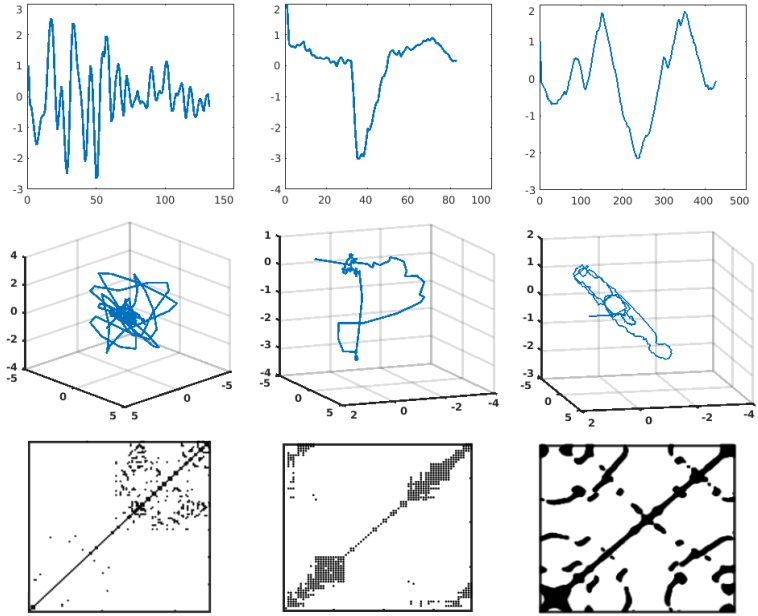}
\caption{Application of RP ($m=3$ and $\tau=4$) on three different datasets from the UCR time-series classification archive \cite{Ref33}: left) FaceAll, middle) TwoLeadECG, and right) Yoga data (from left to right, respectively). First row is the signal plot, second row demonstrates the 3D space trajectory and last row visualize the R-matrix (traditional binary plots using $\theta = 0.5$).}
\label{fig:fig01}
\end{figure*}

\section{Methodology}

Time-series can be characterized by a distinct recurrent behavior such as periodicities and irregular cyclicities. Additionally, the recurrence of states is a typical phenomena for dynamic non-linear systems or stochastic processes that time-series are generated in \cite{Rohde2008,Marwan2009,Thiel2003}. The RP is a visualization tool that aiming to explore the m-dimensional phase space trajectory through a two-dimensional representation of its recurrences. The main idea is to reveal in which points some trajectories return to a previous state and it can be formulated as:

\begin{equation}
\label{eq1}
R_{\rm i,j}=\theta(\epsilon-\Vert \vec{s}_{i} - \vec{s}_{j} \Vert), \hspace{2mm} \vec{s}(.) \in \mathfrak R^{m},  \hspace{2mm} i,j=1,...,K
\end{equation}

where $R_{\rm i,j}$ are elements of the RP matrix, $m$ is dimension of the phase space trajectory, $N$ is the number of considered states $\vec{s}$, $\epsilon$ is a threshold distance, $\Vert . \Vert$ a norm and $\theta(.)$ the Heaviside function (a.k.a the unit step function whose value is zero for negative argument and one for positive argument). There is also another important parameter involve in generating the RP matrix: embedding time delay $\tau$ that can be considered as sub-sampling factor of the time-series signal. The R-matrix contains both texture which are single dots, diagonal lines as well as vertical and horizontal lines; and typology information which characterized as homogeneous, periodic, drift and disrupted. For instance, fading to the upper left and lower right corners means the process contains a trend or drift; or vertical and horizontal lines/clusters show that some states do not change or change slowly for some time and this can be interpreted as laminar states \cite{Ref17}. Application of RP (using the phase space dimension $m=3$ and embedding
time delay $\tau=4$ parameters) on three different exemplar datasets from the UCR time-series classification archive \cite{Ref33} e.g. FaceAll, TwoLeadECG and Yoga data is shown in Figure\ref{fig:fig01}.

Inspired by the unique texture images obtained from the R-matrices, this paper proposes a time-series classification pipeline based on the BoF model. First the raw 1D time-series signals $x^{n}$ are transformed into 2D recurrence texture images, and then the local patches are extracted using blob detectors (here, we applied multi-scale dense grid sampling). The SIFT is a robust and popular computer vision algorithm to detect and describe local features in images. In the proposed BoR model, SIFT operator is applied to describe $K$ patches of an image $S=[s_{1}, s_{2}, .., s_{K}] \in \mathfrak R^{128 \times K}$ (since there are 4 x 4 = 16 histograms each with 8 bins the vector has 128 elements). Then, descriptors are used as the input for the k-means clustering algorithm in order to generate the \emph{initial} dictionary of length $M$, $D_{init}=[w_{1}, .., w_{M}]$, where $w_{i}$ is the $i$th initial visual word and optimum $M$ is determined experimentally and depends on the problem. The k-means clustering aims to partition samples into $M$ clusters in which each sample belongs to the cluster with the nearest mean, serving as a prototype of the cluster.

\begin{figure*}[t!]
\centering
\includegraphics[scale=0.68]{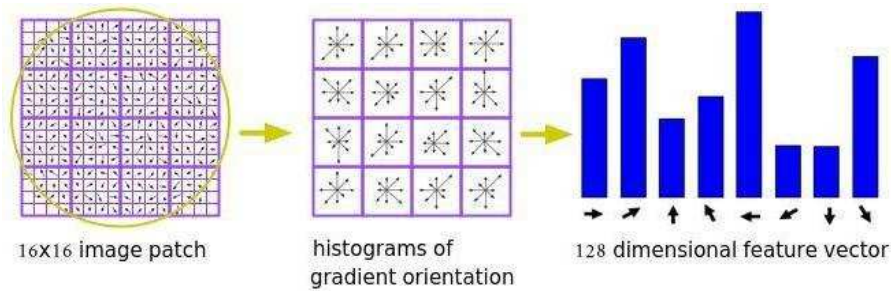}
\caption{The SIFT descriptor generation. First the gradient magnitude and orientation at each image sample point in a $16 \times 16$ patch is computed (left). These samples are then accumulated into orientation histograms summarizing the contents over $4 \times 4$ sub regions. The length of each arrow corresponding to the sum of the gradient magnitudes near that direction within the region (middle). The final 128 ($16 \times 8$) dimensional feature vector is obtained by concatenating the histograms (right).}
\label{fig:sift}
\end{figure*}

There is a big effort in the BoF and dictionary learning literature for improving the quality of the codebooks. The proposed BoR model uses an online incremental codebook optimization method, proposed by \cite{Ref32}, based on the Locality-constrained Linear Coding (LLC). Instead of the traditional BoF image representation using histogram of words, the main idea of the LLC is to project each descriptor into its local-coordinate system, followed by max-pooling combined with $\ell^{2}$ normalization as suggested in \cite{Ref41}, to generate the feature vectors. LLC applies locality constraint to select similar basis of local image descriptors from a codebook, and learns a linear combination weight of these basis to reconstruct each descriptor. For ease of reference, the codebook optimization process is shown in Algorithm \ref{alg:alg01}. After optimizing the initial dictionary, $D_{opt}=[w^{*}_{1}, .., w^{*}_{M}]$, the coding process is done in 3 steps: i) finding k-nearest words of SIFT descriptor $s_{i}$ denoted as $D_{i}$, ii) reconstructing $s_{i}$ using $D_{i}$ by $\tilde{c} = argmax_{c} \Vert s_{i} - D_{i} c_{i}\Vert ^{2},\hspace{2mm} s.t. \sum^k_j c_{j}=1$, which $c_{i}$ is a code for $S$. One of the advantages of the LLC coding is that classification of its representation (features) using linear classifier obtains better results compared to the traditional spatial pyramid matching (SPM) using non-linear classifier \cite{Ref32}. The readers are referred to \cite{Ref32} for more details on the LLC. It worth noting that according to our experimental results (see the next section), the codebook generated by K-Means can obtain satisfactory results. However in this work, the LLC coding criteria is used to train the codebook, which further improves the accuracy.

\begin{algorithm}
\caption{LLC Incremental codebook optimization \cite{Ref32}.}
\label{alg:alg01}
\bf{input:} $D_{init}\in \mathfrak S^{128\times M}$, $S \in \mathfrak R^{128 \times K}$, $\mu$, $\sigma$, $\hspace{3mm}$ \bf{output:} $D_{opt}$

\begin{algorithmic}[1]

\State $D_{opt} \leftarrow D_{init}$
\For {$i = 1:N$}
\State $d \leftarrow 1\times M $ $zero$ $vector,$
\For {$j = 1:M$} $\hspace{3mm}$ // locality constraint parameters
\State $d_{j} \leftarrow exp^{-1}(-\Vert s_{i}-b_{j} \Vert ^{2}/ \sigma)$
\EndFor
\State \bf{end for}
\State $d \leftarrow normalize_{(0,1]} (d)$
\State $ c_{i} \leftarrow argmax_{c} \Vert s_{i} - Dc \Vert^{2} + \mu 
\Vert d \odot c \Vert ^{2},\hspace{3mm} s.t. \hspace{1mm} 1^{T} c = 1$ // coding
\State $id \leftarrow {j \mid abs(c_{i}(j)) > 0.01}, D_{i}  \leftarrow D(:,id)$  // remove bias
\State $\tilde{c}_{i} \leftarrow argmax_{c} \Vert s_{i} - D_{i} c\Vert ^{2},\hspace{3mm} s.t. \hspace{1mm} \sum_j c(j) = 1$
\State $\triangle D_{i} \leftarrow -2\tilde{c}_{i}(s_{i} - D_{i} \tilde{c}_{i}), \hspace{3mm}\mu \leftarrow \sqrt{1/i}$ $\hspace{2mm}$ // update basis
\State $D_{i} \leftarrow D_{i} - \mu \triangle D_{i} / \mid \tilde{c}_{i} \mid_{2}$
\State $D_{opt}(:,id) \leftarrow proj(D_{D_{i}})$
\EndFor
\State \bf{end for}
\end{algorithmic}
\end{algorithm}

\begin{figure*}[t!]
\centering
\includegraphics[scale=.38]{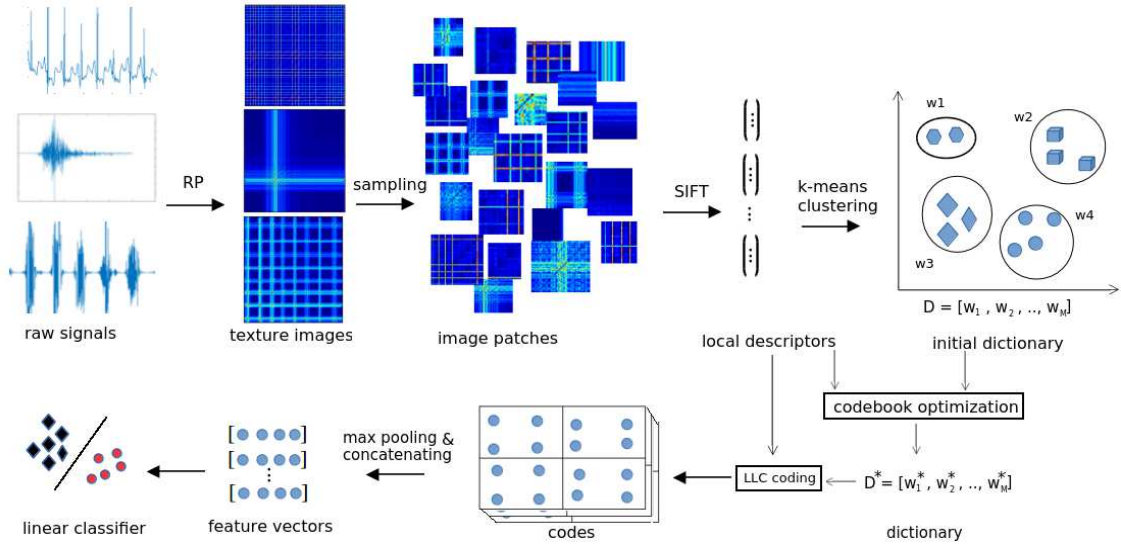}
\caption{The proposed BoR pipeline for time-series classification.}
\label{fig:figure02}
\end{figure*}

The final feature vectors of time-series signals are classified using a linear SVM classifier. It is worth noting that the resulting R-matrix by Formula (1) has only \{0,1\} values, that caused by thresholding parameter $\epsilon$. In order to avoid information loss by binarization of the R-matrix, this paper skips the thresholding step and uses the gray-level texture images. The proposed BoR pipeline for time-series classification is shown in Figure \ref{fig:figure02}.\\
\indent The proposed BoR model opens a new window of opportunities for time-series classification by treating the 1D signal classification as the computer vision problem of texture recognition. Note that the proposed pipeline is an exemplar case. Different features other than SIFT such as HOG \cite{Ref44} and LBP \cite{Ref40} can be extracted (or combination of these descriptors) in order to better represent the signals, that hopefully lead to a better accuracy of the over-all system. 

\section{Experimental results and discussions}
\begin{figure}
   \begin{center}
   \begin{tabular}{c}
   \includegraphics[scale=.5]{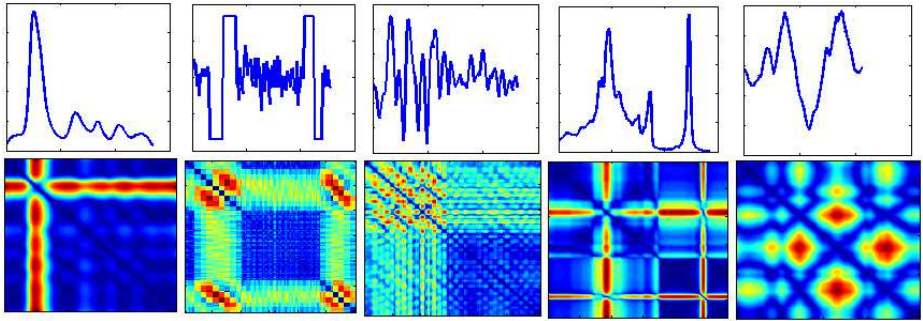}
   \end{tabular}
   \end{center}
   \caption{Application of RP ($m=3, \tau = 4$) time-series to image encoding on five different datasets from the UCR archive: 50words, TwoPatterns, FaceAll, OliveOil and Yoga data (from left to right, respectively) without binarization.} 
\label{fig:RP_UCR} 
\end{figure}

In order to evaluate the performance of the proposed method, the UCR time-series classification archive is used \cite{Ref33}. For having a comprehensive evaluation of the classification algorithms, this repository contains 85 time-series datasets with different characteristics i.e. the number of classes $2 \leq N_{c} \leq60$, number of training samples $16 \leq N_{tr} \leq 8926$ and time-series length $24 \leq l \leq 2709$. The datasets are obtained from varieties of different real-world applications, ranging from EEG signals to food/beverage flavors, and electric devices to phonemes signals. We have selected the same 45 datasets that the results of the other state-the-art algorithms are also reported for them. The training and testing sets are separated to make sure that the results of different algorithms of different studies are comparable. Furthermore, the error rates of "1-NN Euclidean Distance", "1-NN DTW(r) where r is the percentage of time series length”, and "1-NN DTW (no Warping Window)" are reported as the base-line methods. Only the results of 1-NN DTW is included in the table, because it obtains the lower recognition rate compared to the other two \cite{Ref33}. For a better comparison, 45 datasets that the results of the majority of related algorithms exist also for them are chosen (see Table \ref{tab1}). Application of RP (with the phase space dimension $m=3$, and embedding time delay $\tau = 4$) time-series to image encoding on first sample of five different datasets from the UCR archive are shown in Figure \ref{fig:RP_UCR}.

\begin{table}
\begin{center}
\begin{tabular}{|l|c|c|c|c|c|c|c|}
\hline
Parameters & patch size & codebook size & bag size & $C$ (SVM) & $m$ & $\tau$\\
\hline
Values & \{ 16,24,32,48\} & \{ 50,100,250,...,8k\} & $>$10,000 & $2^{-10}$-$2^{10}$ & \{3,4,5\} & \{1,3,10\}\\
\hline
\end{tabular}
\end{center}
\caption{Different parameters of the proposed BoR model and their value ranges.}
\label{tab:parameters}
\end{table}

\begin{table*}[!t]
\caption{\label{tab1}Comparison of the proposed BoR method and the state-of-the art algorithms on the selected UCR datasets in terms of error rates (\%).}
\centering
\small
\begin{tabular}{|p{1.2cm}|p{0.8cm}|p{0.8cm}|p{0.8cm}|p{1cm}|p{1cm}|p{1cm}|p{1.3cm}|p{1.4cm}|p{1cm}|}

\hline

Dataset & $N_{c}$ & $N_{tr}$ & $N_{te}$ & $l$ & 1-NN\newline DTW 

& SVM-SAX& BoTSW\cite{Ref25} & BoF \cite{Ref18} & proposed\newline BoR\\

\hline
50words & 50 & 450 & 455 & 270  & 24.2& 37.4& 36.3& \bf{20.9}& 37.1\\

\hline
Adiac & 37 & 390 & 391  & 176& 39.1& 36.6& 61.4 & 24.5 &\bf{23.0}\\

\hline
Beef & 5 & 30 & 30  & 470& 46.7& 40.0& 40.0& 28.7&\bf{26.6}\\


\hline
CBF & 3 & 30 & 900  & 128& \bf{0.4}& 4.8 & 5.8& 0.9 &1.8\\

\hline
Chlori. & 3 & 467 & 3840  & 166& 35& 34.2& --& 33.6& \bf{33.2}\\

\hline
Cin.ECG & 4 & 40 & 1380  & 1639& \bf{6.4}& 10.4& --& 26.2 &25.5\\

\hline
Coffee & 2 & 28 & 28  & 286& 17.9& 14.3 & \bf{0}& 0.4 &\bf{0}\\

\hline
CricketX & 12 & 390 & 390  & 300& 23.6& 42.6 & --& 27.8 &\bf{0}\\

\hline
CricketY & 12 & 390 & 390  & 300& \bf{19.7}& 34.6& --& 25.9 &27.4\\

\hline
CricketZ & 12 & 390 & 390  & 300& \bf{17.9}& 39 & --& 26.3 &23.7\\

\hline
Diatom. & 4 & 16 & 306  & 345& \bf{6.5} & 14.7& --& 12.6 &9.8\\

\hline
ECG200 & 2 & 100 & 100  & 96& 12.0& 18.0& 11.0& 14.5 & \bf{10.8}\\

\hline
ECG5day & 2 & 23 & 861  & 136& 20.3& \bf{14.0} & --& 18.3 &16.9\\

\hline
FaceAll & 14 & 560 & 1690  & 131& 19.2& 24.6& 21.8& 23.4 & \bf{15.7}\\

\hline
Face4 & 4 & 24 & 88  & 350 & 11.4& \bf{0}& \bf{0}& 5.1& 9.0\\

\hline
Fac.UCR & 14 & 200 & 2050  & 131& \bf{8.8}& 16.0& --& 9.0 &21.7\\

\hline
Fish & 7 & 175 & 175  & 463& 16.0& 5.7& 6.9& 8.0 &\bf{2.8}\\

\hline
GunPoi. & 2 & 50 & 150  & 150& 8.7& 2& 8.0& 1.1 &\bf{0}\\

\hline
Haptics & 5 & 155 & 308  & 1092& 58.8& 56.2& --& 48.8 &\bf{46.2}\\

\hline
Inline. & 7 & 100 & 550  & 1882&  \bf{61.1} & 88.5& --& 60.3 & 61.4\\

\hline
Ita.Pa. & 2 & 67 & 1029  & 24&  \bf{4.5}& 9.2& --& 9.6 & 5.8\\

\hline
Light.2 & 2 & 60 & 61  & 637&  \bf{13.1}& 24.6& 36.1& 25.7 &20.3\\

\hline
Light.7 & 7 & 70 & 73  & 319 &  28.8& 39.7& 38.4& 26.2 & \bf{25.5}\\

\hline
MALLAT & 8 & 55 & 2345  & 1024& 8.6& 29.3& --& 3.7 &\bf{3.4}\\

\hline
Med.Im. & 10 & 381 & 760  & 99& \bf{25.3}& 47.4& --& 26.9 & 34.8\\

\hline
Mot.St. & 2 & 20 & 1252  & 84& \bf{13.4} & 50.6& --& 13.5 & 19.0\\



\hline
OliveO. & 4 & 30 & 30  & 570& 16.7& 13.3& 10.0& \bf{9.0} & 20.8\\

\hline
OSULea. & 6 & 200 & 242  & 427& 38.4& 22.3& 18.2& 32.9 & \bf{8.1}\\

\hline
Son.Ro. & 2 & 20 & 601  & 70& 30.4& 16.3 & --& 17.5 & \bf{16.0}\\

\hline
So.RII & 2 & 27 & 953  & 65& \bf{14.1} & 27.3& --& 19.6 & 26.2\\

\hline
StarCu. & 3 & 1000 & 8236  & 1024& 9.5& 4.2& --& \bf{2.2} & 3.9\\

\hline
SwedLe. & 15 & 500 & 625  & 128& 15.5& 14.4& 15.2& 7.5 &\bf{6.8}\\

\hline
Symbols & 6 & 25 & 995  & 398& 6.2& 5.7& --& \bf{3.4} & 4.8\\

\hline
SyntCo. & 6 & 300 & 300  & 60& 1.7& 1.3& 4.3& 0.8 &\bf{0.6}\\

\hline
Thorax1 & 42 & 1800 & 1965  & 750& 18.5& -- & --& 13.8 & \bf{13.5}\\

\hline
Thorax2 & 42 & 1800 & 1965  & 750& 12.9& --& --& 13.0 & \bf{12.7}\\

\hline
Trace & 4 & 100 & 100  & 275& 1.0& \bf{0} & 1.0& 2.0 & \bf{0}\\

\hline
TwoLead & 2 & 23 & 1139  & 82& 11.6& 7.6& --& 4.6 & \bf{3.1}\\

\hline
TwoPat. & 4 & 1000 & 4000  & 128& 2.0& 7.0& 0.2 & 0.1 & \bf{0}\\

\hline
uWaveX & 8 & 896 & 3582  & 315& 22.7 & 76.5& --& \bf{16.4} & 21.9\\

\hline
uWaveY & 8 & 896 & 3582  & 315& 30.1& 37.3& --& \bf{24.9} & 29.5\\

\hline
uWaveZ & 8 & 896 & 3582  & 315& 32.2& 33.5& --& 21.7 & \bf{20.2}\\

\hline
Wafer & 2 & 1000 & 6174  & 152& 5.0& 9.0& 0.1& 0.4 & \bf{0}\\

\hline
Wor.Syn. & 25 & 267 & 368  & 270& \bf{25.2} & 45.8& --& 30.2 & 63.7\\

\hline
Yoga & 2 & 300 & 3000  & 426& 15.5& 16.1& 15.0& 14.9 & \bf{13.7}\\

\hline
\hline
\bf{\# wins} & -- & -- & --  & -- & 13/45& 3/43& 2/20& 6/45 & \bf{24/45}\\

\hline
\bf{AveRank} & -- & -- & --  & -- & 2.7& 3.3 & 3.2 & 2.2 &\bf{1.8}\\

\hline
\end{tabular}
\end{table*}

The obtained results are also compared to the state-of-the art BoF algorithms applied on time-series signals. One of the main work in this direction is time-series classification with a bag-of-features (TSBF) by \cite{Ref18}. Multiple random and uniform subsequences are sampled for feature extraction (e.g. slopes, means, and variances) and labeled with the class of the time-series. A random forest (RF) ensemble is used to both calculates "class probability estimates" (CPEs) for codebook and histogram of the CPEs to summarize the subsequence information. After adding the global features, a final classifier is used to assign each time series. Bag-of-Temporal-SIFT-Words (BoTSW) \cite{Ref34,Ref25} is the other algorithm that appears in the table. BoTSW tries to adapt SIFT image descriptors to time-series where the words correspond to the description of local gradients, computed at different scales, around keypoints. 

For the comparison purposes, "Number of Wins" and "Average Rank" are calculated for each algorithm. Number of Wins counts number of datasets that a specific algorithm obtains the lowest error rates, while Average Rank is the mean of the error-rate ranking over entire datasets. Since some error rates are missing for some datasets for some algorithms, we normalized the measures for each algorithm. An algorithm outperforms the rest if it has a highest number of wins and lowest average rank.

For the computational efficiency, gray-level images with big dimensions resized down to maximum dimension of 300 pixels. In order to obtain image patches, a multi-scale Dense-SIFT method is used, with the different local image patch size, under 4 following scales: 16$\times$16, 24$\times$24, 32$\times$32 and 48$\times$48. The optimum size of the codebook, $M$, is one of the most important parameters in BoF that has to be selected experimentally. Too small $M$ has a limited discriminative ability, while a large size is likely to introduce noise. We used 5-fold cross-validation (CV) to select $M$ from the set $\{50, 100, 250, 500, 1000, 2000, 4000, 8000\}$, with maximum number of k-means iterations set to 50. Regarding the LLC parameters, the number of neighbours was set to 5 with shift-invariant constraint. For assigning class labels, LIBSVM \cite{Ref35} is used as a linear classifier. The best $C$ parameter of SVM was determined based on CV by varying $C$ between 10 logarithmically equal spaced points between $2^{-10}$ and $2^{10}$. Binary classifiers are extended by 1-vs-all method to address multi-class decisions (although techniques such as Error-Correcting Output Codes -ECOC- could have achieved a better performance \cite{Ref36,Ref37,Ref38,Ref42} and we did not investigate it in this work). 

\begin{figure}[htp]
\centering
\includegraphics[width=.65\textwidth]{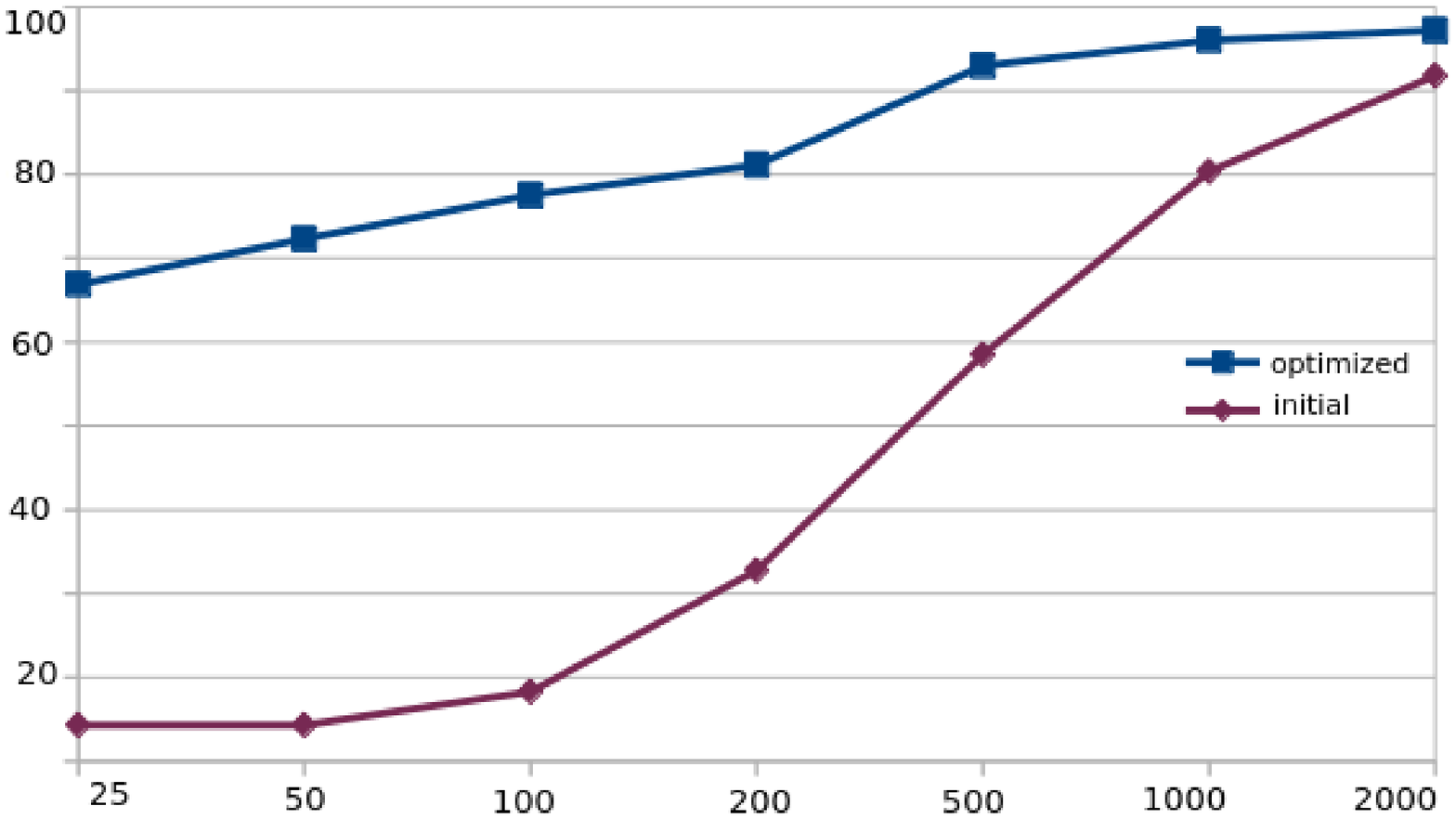}
\includegraphics[width=.65\textwidth]{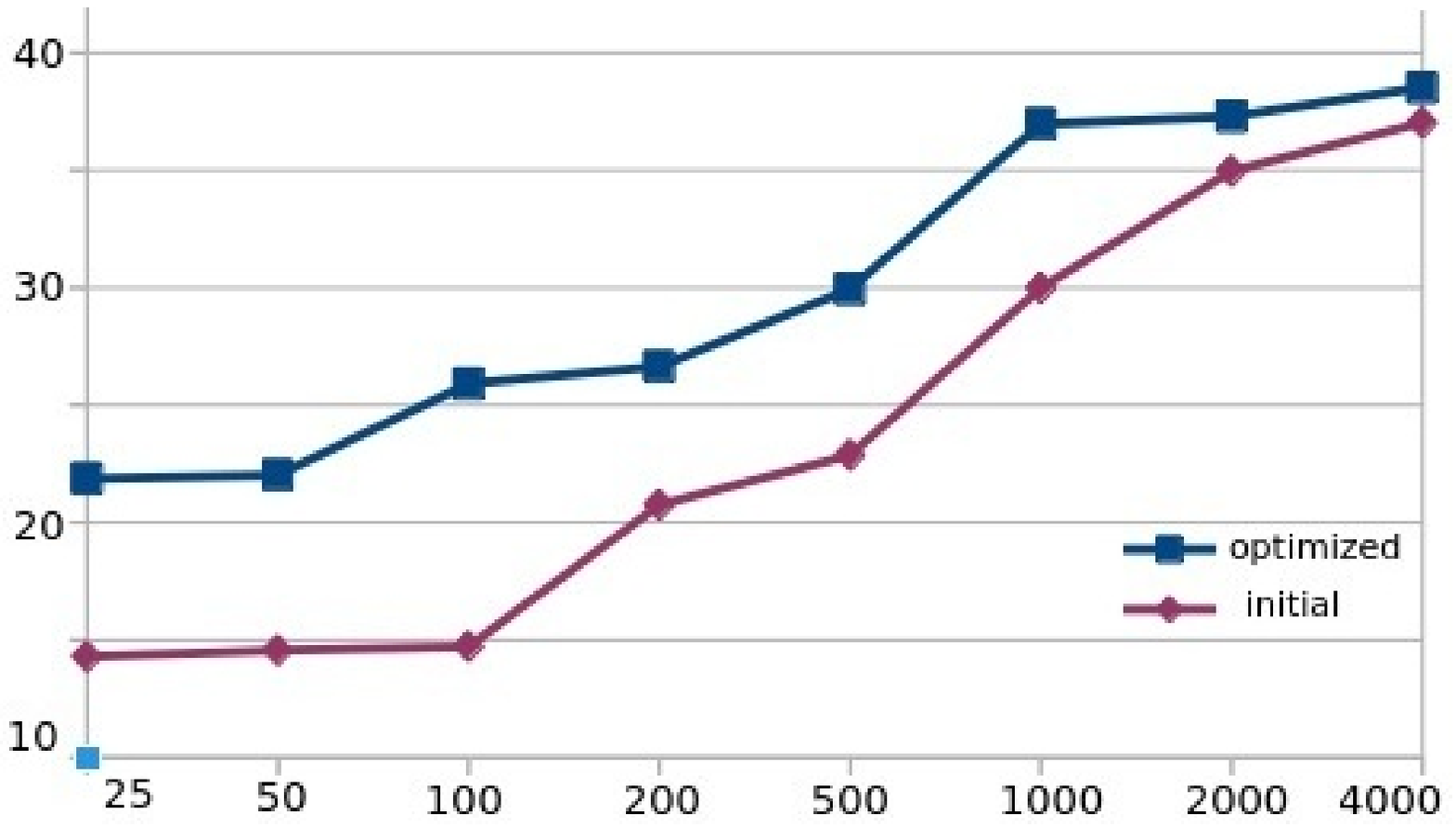}
\caption{Performance (in terms of recognition rates) of the initial dictionaries vs. optimized dictionaries  over different codebook sizes for two arbitrarily chosen datasets (top:fish; bottom:inlineSkate datasets).}
\label{fig:figure04}
\end{figure}

Performances of the initial dictionary compared to the optimized one obtained by the LLC-based algorithm \ref{alg:alg01} for two arbitrarily chosen datasets are shown in Figure \ref{fig:figure04}. It demonstrates the improvement of the input dictionary, regardless of the codebook size. However, the experiments show that the improvements are more significant for smaller size of codebooks and greater dictionary sizes benefit less from the optimization process.

According to the Table \ref{tab1}, the proposed method obtained the lowest average ranking (1.8) and greater number of wins (24 out of 45). The second most accurate algorithm, based on the average rank, is the BoF proposed by \cite{Ref18} that uses bag of signal segments in different scales and also takes advantage of the classifier ensemble (average rank = 2.2). However, if the number of wins is taken into account, 1NN DTW with 13 out of 45 is the third accurate algorithm. The results also demonstrate that although the 1NN DTW is a simple classifier, still achieves a better position compared to the popular SAX and BoTSW algorithms. The results also demonstrate that the transforming the time-series signal into 2D using RP in the BoF framework boosts the accuracy of the system.  

\begin{figure}[htp]
\centering
\includegraphics[width=.65\textwidth]{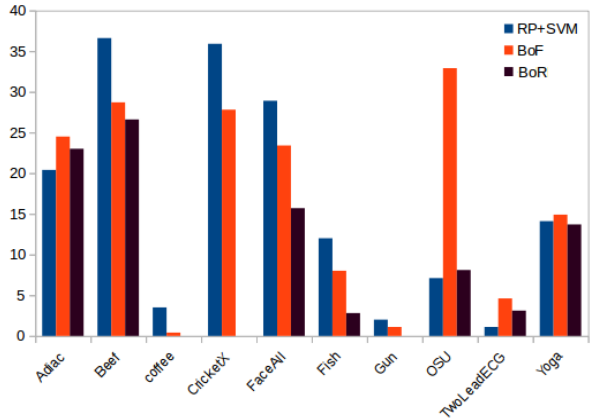}
\caption{Comparison (in terms of the error rate percentage) between BoF \cite{Ref18}, RP+SVM \cite{RP_svm} and their combination (BoR) on 10 datasets from the UCR archive.}
\label{fig:figure05}
\end{figure}

It is known from the previous work that RP can boost time-series classification performance \cite{RP_svm,hatami-cnn}. Furthermore, recently BoF classifiers widely and successfully applied for TSC tasks \cite{Ref18}. Figure \ref{fig:figure05} compares the performance of these two approaches against the proposed idea, which is somehow combination of them. Some experiments have been carried out in order to investigate which idea were responsible for most of the improvements, or is the combination itself that is uniquely useful. For 10 arbitrarily chosen datasets from the UCR archive, the proposed BoR obtains lower average error rate. 

\begin{figure*}[t!]
\centering
\includegraphics[scale=0.65]{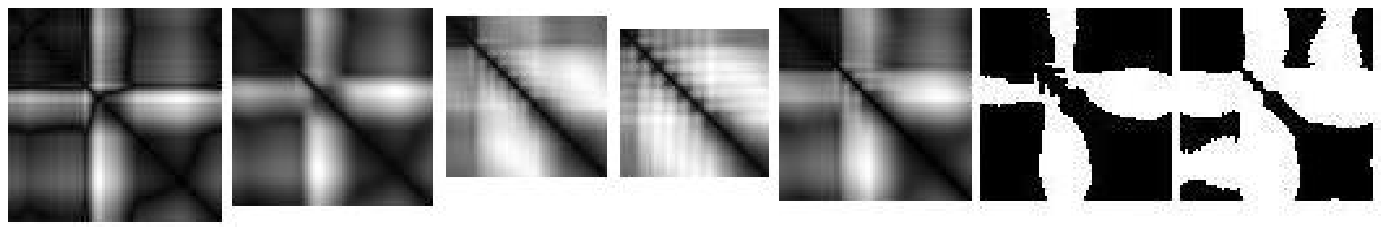}
\includegraphics[scale=0.65]{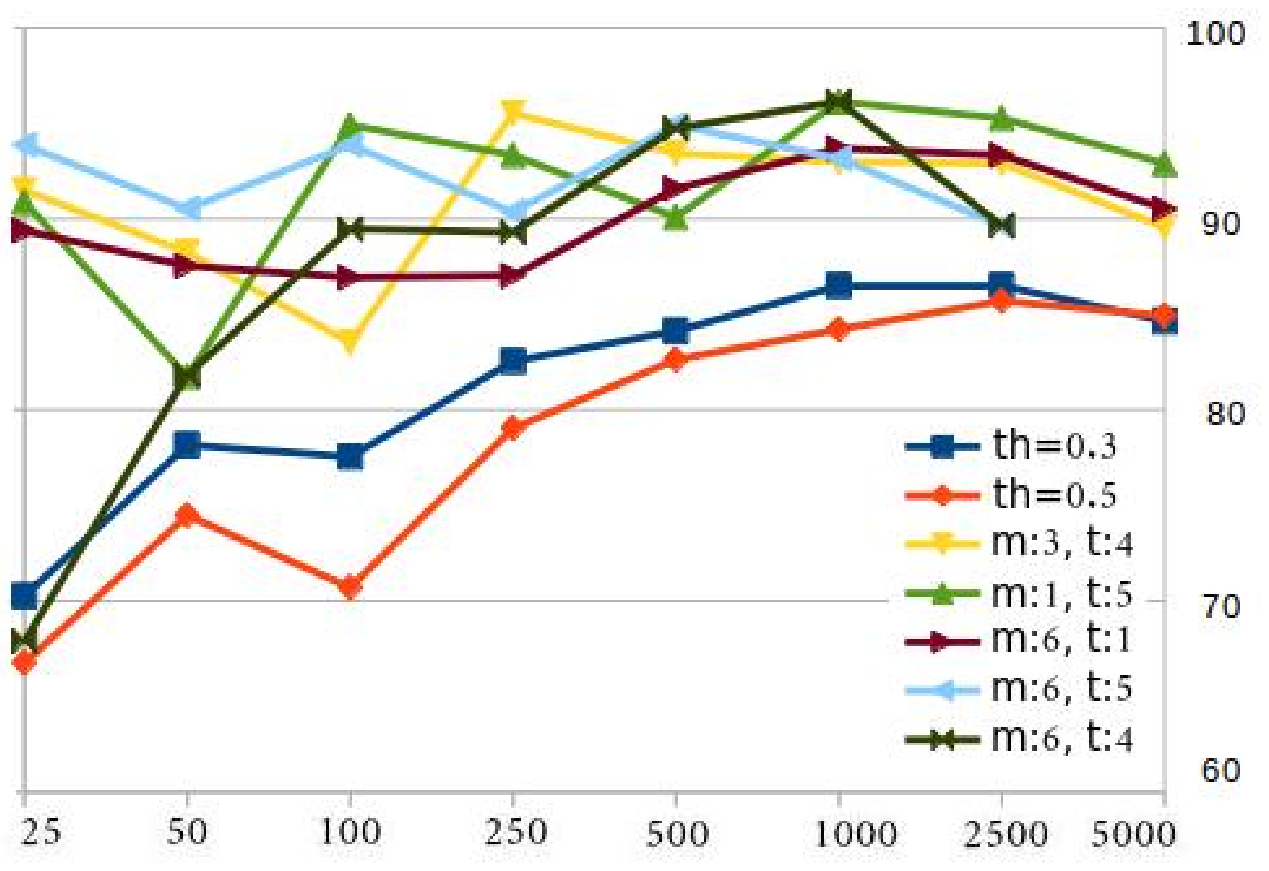}
\caption{Comparisons of different RP parameters on TwoLeadECG data of the UCR archive.
Top: RP of first sample using (m:1, $\tau$:5), (m:6, $\tau$:1), (m:6, $\tau$:4), (m:6, $\tau$:5), (m:3, $\tau$:4), (m:3, $\tau$:4, $th$:0.3), and (m:3, $\tau$:4, $th$:0.5) from left to right, respectively; Bottom: performances (in terms of accuracy) of BoR model using different parameters.}
\label{fig:RP_params}
\end{figure*}

Figure \ref{fig:RP_params} compares the performance of the proposed BoR model using different RP parameters e.g. $m$ and $\tau$. Furthermore, we also investigate the impact of using gray-level images instead of the traditional binary images for TSC. The TwoLeadECG data from the UCR archive is chosen to run this experiment. Application of the RP on one sample using different parameters e.g. (m:1, $\tau$:5), (m:6, $\tau$:1), (m:6, $\tau$:4), (m:6, $\tau$:5), (m:3, $\tau$:4), (m:3, $\tau$:4, $th$:0.3), and (m:3, $\tau$:4, $th$:0.5) is shown on the top of the Figure \ref{fig:RP_params} from left to right, respectively. On the bottom, performances (in terms of accuracy) of the BoR model using different parameters is reported on the TwoLeadECG data. As visually noticeable, binarization of the RP images looses texture information (Fig. \ref{fig:RP_params} top) and therefore, the lower performance of the BoR model for binary RPs compared to the gray-level images are explained (Fig. \ref{fig:RP_params} bottom). Regarding the sensitivity analysis of the main parameters ($m$, $\tau$ and codebook size $M$) of the BoR model, as shown, for a given ($m$, $\tau$) there is a suitable $M$ that obtains a maximum performance ($\sim$ 93\%, in this case).

\begin{figure*}[t!]
\centering
\includegraphics[scale=0.6]{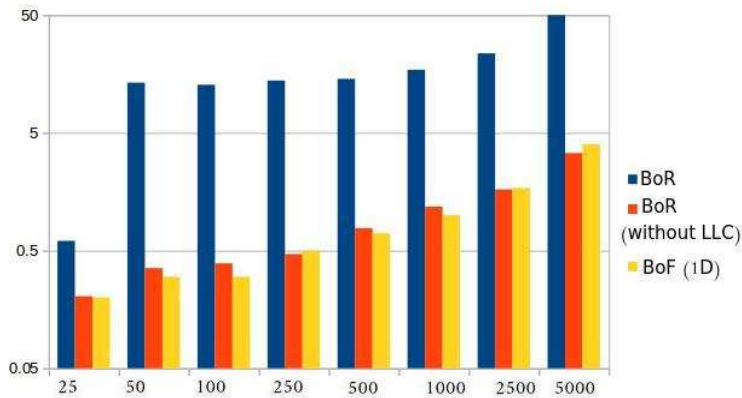}
\caption{Comparison of run-time (in seconds) of the proposed BoR with the traditional 1D BoF model on the TwoLeadECG dataset.}
\label{fig:time_cost}
\end{figure*}

Figure \ref{fig:time_cost} compares the computational complexity (in terms of run-time) of the proposed BoR with the traditional 1D BoF model on the TwoLeadECG dataset. As shown, the LLC module of the proposed approach is consuming a significant time, in exchange for the accuracy boost (particularly for small $M$s, see Figure \ref{fig:figure04}). However, the proposed BoR model without the LLC module has a similar run-time and also obtains a higher accuracy, compared to the traditional BoF model (usually requires a bigger $M$ compared to the proposed BoR - see Figure \ref{fig:figure04}). 

\section{Conclusions and future directions}

This paper proposed image representation of time-series signals in BoF framework. By treating a time-series classification problem as a texture recognition task, this strategy facilitates application of new features from computer vision and image processing domains for time-series signals. This allows extracting more informative and discriminative information that can boost the recognition rate. The experimental results and analysis conclude that the proposed BoR approach has competitive performance when comparing to the 1D BoF applied on time-series. This suggests that for BoF approach, more discriminative and informative words can be obtained in 2D space compared to the words obtained from signal segments. 


In this study, we used an existing time series-to-image transform i.e. RP for generating texture images. The future work will be devoted to finding/proposing a novel and more representative time-series to image transform. Furthermore, exploring different image descriptors such as LBP, gist, SURF and HoG, and also combination of them with SIFT may lead to a more discriminative dictionary that need to be investigated. Finally, performance of a hybrid method by combining words from both 1D segments and image representation of the words may also enrich the dictionary \cite{Ref43}. Application of the other classification algorithms such as deep neural networks is also worth investigating for boosting the recognition rate.

\begin{acknowledgements}

This research is partially supported by the French national research agency (ANR) under the PANDORE grant with reference number ANR-14-CE28-0027.

N. Hatami also acknowledges support of the LABEX PRIMES (ANR-11-LABX-0063) of Universit\'e de Lyon, within the program "Investissements d'Avenir" (ANR-11-IDEX-0007).
\end{acknowledgements}


\begin{thebibliography}{}

\bibitem{Acharya2011}
Acharya, U., Sree, S., Chattopadhyay, S., Yu, W., Ang, P., Application of recurrence quantification analysis for the automated identification of epileptic EEG signals, International journal of neural systems 21 (03), 199-211, 2011.

\bibitem{Allili_2012}
Allili, M., Wavelet modeling using finite mixtures of generalized Gaussian distributions: application to texture discrimination and retrieval, IEEE Transactions on Image Processing 21 (4), 1452-1464, 2012.

\bibitem{Ref37}
Armano, G., Chira, C., Hatami, N., 2012. Error-correcting output codes for
multi-label text categorization. 3rd Italian Information Retrieval Workshop
(IIR), 26-37.

\bibitem{Ref25} 
Bailly, A., Malinowski, S., Tavenard, R., Guyet, T., Chapel, L., 2015. Bag-of-temporal-sift-words for time series classification. ECML/PKDD Workshop
on Advanced Analytics and Learning on Temporal Data .

\bibitem{Ref34} 
Bailly, A., Malinowski, S., Tavenard, R., Guyet, T., Chapel, L., 2016. Dense
bag-of-temporal-sift-words for time series classification. Lecture Notes in
Artificial Intelligence.

\bibitem{Ref49} 
Baydogan, M., Runger, G., 2016. Time series representation and similarity
based on local auto patterns. Data Mining and Knowledge Discovery 30(2).

\bibitem{Ref18} 
Baydogan, M., Runger, G., Tuv, E., 2013. A bag-of-features framework to
classify time series. IEEE Trans. on PAMI 35(11), 2796-2802.

\bibitem{Ref23} 
Bromuri, S., Zufferey, D., Hennebert, J., Schumacher, M., 2014. Multi-label
classification of chronically ill patients with bag of words and supervised
dimensionality reduction algorithms. Journal of biomedical informatics 51,
165-175.

\bibitem{Ref35} 
Chang, C., Lin, C., 2011. Libsvm: a library for support vector machines. ACM
Trans. on Intelligent Systems and Technology 2(3), 27.


\bibitem{Ref33} 
Chen, Y., Keogh, E., Hu, B., Begum, N., Bagnall, A., Mueen, A., Batista, G., 2015. The ucr time series classification archive.

\bibitem{Chen_2012}
Chen, Y., Yang, H., Multiscale recurrence analysis of long-term nonlinear and nonstationary time series, Chaos, Solitons \& Fractals 45 (7), 978-987, 2012.

\bibitem{Ref44} 
Dalal, N., Triggs, B., 2005. Histograms of oriented gradients for human detection. CVPR, San Diego.

\bibitem{Dong_2017} 
Dong, Y., Feng, J., Liang, L., Zheng, L., Multiscale Sampling Based Texture Image Classification, IEEE Signal Processing Letters, Vol.24, Issue.5, 614-618, 2017.

\bibitem{Dong_2015} 
Dong, Y., Tao, D., Li, X., Ma, J., Pu, J., Texture classification and retrieval using shearlets and linear regression, IEEE transactions on cybernetics 45 (3), 358-369, 2015.

\bibitem{Ref12} 
Eads, D., Glocer, K., Perkins, S., Theiler, J., 2005. Grammar-guided feature extraction for time series classification. NIPS.

\bibitem{Eckmann} 
Eckmann, JP. Kamphorst, SO. Ruelle, D., Recurrence Plots of Dynamical Systems, Europhysics Letters, 4(9), 973-977. (1987).

\bibitem{Ref28} 
Fu, Z., Lu, G., Ting, K., Zhang, D., 2011. Music classification via the bag-of-features approach. Pattern Recognition Letters 32(14), 1768-1777.

\bibitem{Ref11}
Geurts, P., 2009. Pattern extraction for time series classification. 5th EU Conf. Principles of Data Mining and Knowledge Discovery, 115-127.

\bibitem{Ref22} 
Grabocka, J., Wistuba, M., Schmidt-Thieme, L., 2015. Scalable classification
of repetitive time series through frequencies of local polynomials. IEEE
Trans. on Knowledge and Data Engineering 27(6), 1683-1695.

\bibitem{hatami-cnn}
Hatami, N., Gavet, Y., Debayle, J., Classification of Time-Series Images Using Deep Convolutional Neural Networks, International Conference on Machine Vision (ICMV), 2017.

\bibitem{Ref42} 
Hatami, N., 2008. Thinned ecoc decomposition for gene expression based cancer classification. 8th IEEE Conf. on Intelligent Systems Design and Applications, 216-221.

\bibitem{Ref43} 
Hatami, N., 2012a. Some proposals for combining ensemble classifiers. PhD
thesis, University of Cagliari.

\bibitem{Ref36} 
Hatami, N., 2012b. Thinned-ecoc ensemble based on sequential code shrinking.
Expert Systems with Applications 39(1), 936-947.

\bibitem{Ref5} 
Hatami, N., Chira, C., 2013. Classifiers with a reject option for early time-series classification. IEEE Sym. on Computational Intelligence and Ensemble Learning (CIEL), 9-16.

\bibitem{Ref38} 
Hatami, N., Ebrahimpour, R., Ghaderi, R., 2008. Ecoc-based training of neural
networks for face recognition. IEEE Conf. on Cybernetics and Intelligent
Systems, 450-454.

\bibitem{Ref6} 
Jeong, Y., Jeong, M., Omitaomu, O., 2011. Weighted dynamic time warping
for time series classification. Pattern Recognition 44(9), 2231-2240.

\bibitem{Ref7} 
Keogh, E., Kasetty, S., 2003. On the need for time series data mining benchmarks: A survey and empirical demonstration. Data Mining and Knowledge
Discovery 7(4), 349-371.

\bibitem{vis_ts}
N. Kumar, V. Lolla, E. Keogh, S. Lonardi, C. Ratanamahatana, L. Wei, 2005. Time-series bitmaps: a practical visualization tool for working with large time series databases, SIAM international conference on data mining, 531-535.

\bibitem{Lategahn_2010}
Lategahn, H., Gross, S., Stehle, T., Aach, T., Texture classification by modeling joint distributions of local patterns with Gaussian mixtures, IEEE Transactions on Image Processing 19 (6), 1548-1557, 2010.

\bibitem{Ref20} 
Lin, J., Khade, R., Li, Y., 2012. Rotation-invariant similarity in time series
using bag-of-patterns representation. Journal of Intelligent Information Systems 39(2), 287-315.

\bibitem{Ref39} 
Lowe, D., 2004. Distinctive image features from scale-invariant keypoints.
International journal of computer vision 60(2), 91-110.

\bibitem{Ref17} 
Marwan, N., Romano, M.C., Thiel, M., Kurths, J., 2007. Recurrence plots for
the analysis of complex systems. Physics Reports 438(5-6), 237-329.

\bibitem{Marwan2002} 
Marwan, N., Wessel, N., Meyerfeldt, U., Schirdewan, A., Kurths, J. (2002). Recurrence Plot Based Measures of Complexity and its Application to Heart Rate Variability Data. Physical Review E. 66 (2): 026702.

\bibitem{Marwan2009} 
Marwan, N. Kurths, J. Comment on “stochastic analysis of recurrence plots with applications to the detection of deterministic signals” by rohde et al.[physica d 237 (2008) 619–629], Physica D 238 (2009) 1711-1715.

\bibitem{Mehta_2016} 
Mehta, R., Eguiazarian, K., Texture classification using dense micro-block difference, IEEE Transactions on Image Processing 25 (4), 1604-1616, 2016.

\bibitem{Ref13} 
Nanopoulos, A., Alcock, R., Manolopoulos, Y., 2001. Feature-based classification of time-series data. Int. J. Computer Research 10, 49-61.

\bibitem{Ref40} 
Ojala, T., Pietikainen, M., Maenpaa, T., 2002. Multiresolution gray-scale and
rotation invariant texture classification with local binary patterns. TPAMI
7(24), 971-987.

\bibitem{Ref8} 
Ratanamahatana, C., Keogh, E., 2004. Making time-series classification more
accurate using learned constraints. Proc. SIAM Intl Conf. Data Mining,
11-22.

\bibitem{Ref16} 
Rodriguez, J., Alonso, C., 2004. Interval and dynamic time warping-based decision trees. ACM Symp. Applied Computing, 548-552.

\bibitem{Ref14} 
Rodriguez, J., Alonso, C., Maestro, J., 2005. Support vector machines of
interval-based features for time series classification. Knowledge-Based Systems 18, 171-178.

\bibitem{Rohde2008} 
Rohde, G., Nichols, J., Dissinger, B., Bucholtz, F., Stochastic analysis of recurrence plots with applications to the detection of deterministic signals, Physica D: Nonlinear Phenomena, 237(5), 619-629, 2008.

\bibitem{Roma2013}
Roma, G., Nogueira, W., Herrera, P., de Boronat, R., Recurrence quantification analysis features for auditory scene classification, IEEE AASP Challenge on Detection and Classification of Acoustic Scenes and Events, 2013.


\bibitem{Ref48}
Schafer, P., 2015. The boss is concerned with time series classification in the
presence of noise. Data Mining and Knowledge Discovery 29(6), 1505-
1530.

\bibitem{Ref21}
Senin, P., Malinchik, S., 2013. Sax-vsm: Interpretable time series classification
using sax and vector space model. Data Mining (ICDM), IEEE 13th Int.
Conf. on, 1175-1180.

\bibitem{silva2013}
Souza, V. Silva, D. Batista, G., Time Series Classification Using Compression Distance of Recurrence Plots, IEEE 13th International Conference on Data Mining, 687-696, 2013.

\bibitem{RP_svm}
Souza, V. Silva, D. Batista, G., 2014. Extracting texture features for time series classification, Pattern Recognition (ICPR), 22nd International Conference on, 1425-1430.

\bibitem{Thiel2003}
Thiel, M. Romano, MC. Kurths, J. Analytical description of recurrence plots of white noise and chaotic processes, arXiv:nlin/0301027, 2003.

\bibitem{Ref9}
Ueno, K., Xi, X., Keogh, E., Lee, D., 2007. Anytime classification using the
nearest neighbour algorithm with applications to stream mining. IEEE IntlConf. Data Mining, 623-632.

\bibitem{Ref19}
Wang, J., Liu, P., She, M., Nahavandi, S., Kouzani, A., 2013. Bag-of-words
representation for biomedical time series classification. Biomedical Signal
Processing and Control 8(6), 634-644.

\bibitem{Ref32}
Wang, J., Yang, J., Yu, K., Lv, F., Huang, T., Gong, Y., 2010. Locality-constrained linear coding for image classification. CVPR, 3360-3367.

\bibitem{Ref31}
Wang, Z., Oates, T., 2014. Time warping symbolic aggregation approximation
with bag-of-patterns representation for time series classification. Machine
Learning and Applications (ICMLA), 13th Int. Conf. on, 270-275.


\bibitem{Ref30}
Wang, Z., Oates, T., 2015. Pooling sax-bop approaches with boosting to classify
multivariate synchronous physiological time series data. FLAIRS Conference, 335-341.

\bibitem{Ref10}
Xing, Z., Pei, J., Yu, P., 2011. Early prediction on time series: A nearest
neighbor approach. Int. Joint Conf. Artificial Intelligence 2168, 1297-1302.

\bibitem{Ref41}
Yang, J., Yu, K., Gong, Y., Huang, T., 2009. Linear spatial pyramid matching
using sparse coding for image classification. CVPR, 1794-1801.

\bibitem{Yang_2011}
Yang, H., Multiscale recurrence quantification analysis of spatial cardiac vectorcardiogram signals, IEEE Transactions on Biomedical Engineering, Vol.58, No.2, 339-347, 2011.

\bibitem{Zbilut1992}
Zbilut, J.P., Webber Jr., C.L. (1992). Embeddings and delays as derived from quantification of recurrence plots. Physics Letters A. 171 (3-4): 199-203.

\bibitem{Ref24}
Zhang, M., Sawchuk, A., 2012. Motion primitive-based human activity recognition using a bag-of-features approach. ACM SIGHIT sym. on Int. health
informatics (IHI), 631-640.

\bibitem{Ref47}
Zhao, J., Itti, L., 2016. Classifying time series using local descriptors with hybrid sampling. IEEE Trans. on Knowledge and Data Engineering 28,
623-637.

  
\end{thebibliography}


\end{document}